# Tsformer: Time series Transformer for tourism demand forecasting

Siyuan Yi[a], Xing Chen[a,*], Chuanming Tang[b]

[a]*Chengdu University of Technology, Chengdu, 610059, China*
[b] *Key Laboratory of Optical Engineering, Institute of Optics and Electronics, Chinese Academy of Sciences, Chengdu 610200, China*

**Abstract**

AI-based methods have been widely applied to tourism demand forecasting. However, current AI-based methods are short of the ability to process long-term dependency, and most of them lack interpretability. The Transformer used initially for machine translation shows an incredible ability to long-term dependency processing. Based on the Transformer, we proposed a time series Transformer (Tsformer) with Encoder-Decoder architecture for tourism demand forecasting. The proposed Tsformer encodes long-term dependency with encoder, captures short-term dependency with decoder, and simplifies the attention interactions under the premise of highlighting dominant attention through a series of attention masking mechanisms. These improvements make the multi-head attention mechanism process the input sequence according to the time relationship, contributing to better interpretability. What's more, the context processing ability of the Encoder-Decoder architecture allows adopting the calendar of days to be forecasted to enhance the forecasting performance. Experiments conducted on the Jiuzhaigou valley and Siguniang mountain tourism demand datasets with other nine baseline methods indicate that the proposed Tsformer outperformed all baseline models in the short-term and long-term tourism demand forecasting tasks. Moreover, ablation studies demonstrate that the adoption of the calendar of days to be forecasted contributes to the forecasting performance of the proposed Tsformer. For better interpretability, the attention weight matrix visualization is performed. It indicates that the Tsformer concentrates on seasonal features and days close to days to be forecast in short-term forecasting.

## 1. Introduction

In the background of the global economic slowdown, tourism spending keeps growing constantly. From 2009 to 2019, the actual growth of international tourism income(54 %) has exceeded global GDP growth (44%) [1]. According to the report by Calderwood & Soshkin, tourism plays a vital role in the global economy and community. In 2018, the tourism industry contributed 10.4% of the global GDP, as well as related jobs. Tourism is fast-growing, tourism demand also keeps growing [2]. Enterprises and governments start to increase the investment in tourism demand forecasting to make tourism demand forecasting more accurate. Accurate tourism demand forecasting benefits managers from enterprises and governments to formulate more efficient public policies and commercial decisions.

Currently, methods based on the time series model have been widely adopted [3]. With the development of deep learning technology, some AI-based methods have been adopted in research, e.g., ANN、LSTM、SVR、

---
* Corresponding author.
  E-mail address: wfdt8019@126.com(Xing Chen).

CNN [3-5]. Generally, AI-based methods achieve better performance, but AI-based methods are considered to be short of interpretability. In the data source aspect, various variables are adopted as features or exogenous variables by current studies to enhance forecasting performance, e.g., weather, holiday, and search engine index [6, 7].

In the recent AI-based time series forecasting methods, ANN does not possess the structure to process time series [8]. The connection of LSTM formed a cycle and makes it possible for the signal to flow in different directions and can process time series [9]. Nevertheless, LSTM is still facing non-parallel training, gradient vanishing when the LSTM network becomes deeper [10], as well as memory capacity limitation. In forward propagation, LSTM computes by time step. For the sake of insufficient memory capacity of cell state, the long-term information could vanish gradually with the process of computation [11]. CNN concerns more about the local feature, and CNN has excellent ability of local feature extraction. However, for global features, multi-layers of CNN are required to obtain a larger receptive field [12]. In the study by Yan et al., the combination of CNN and LSTM is adopted for air quality forecasting in Beijing [13]. Lu et al. proposed GA-CNN-LSTM for daily tourist flow of scenic spots [5]. CNN-LSTM only enhances the concern of short-term patterns but remains the issue of information vanishing of long-term. Another significant bottleneck of AI-based methods including ANN and LSTM is the lack of interpretability, i.e., researchers are unable to reliably recognize which part of the input sequence contributes to the output of the models [14].

We proposed a time series Transformer (Tsformer) based on Transformer for tourism demand forecasting. The advantages of the proposed Tsformer are parallel computation and long-term dependency processing, and the residual connection allows the Tsformer to stack deeper. The Tsformer overcomes non-parallel training, gradient vanishing and memory capacity limitation of LSTM, and the requirement of deeper CNN to obtain a larger receptive field. Currently, Transformer is widely used in natural language processing (NLP) and started to be adopted in the computer vision field, but few studies introduce Transformer into time series forecasting. Currently, the study using Transformer on tourism demand forecasting has not been performed yet. To exploit the excellent long-term dependency capturing ability of Transformer on tourism demand forecasting, we applied architecture and attention masking mechanism improvements to Transformer and proposed the Tsformer for tourism demand forecasting. Moreover, the calendar of days to be forecasted was adopted to enhance the forecasting performance. In order to compare the performance of the proposed Tsformer and commonly used methods, experiments were designed to compare the Tsformer and nine baseline methods on the Jiuzhaigou valley and Siguniang mountain tourism demand datasets. The results of the experiments indicate that the Tsformer performs better than all baseline methods in both short-term and long-term tourism demand forecasting tasks. Based on the experiments, ablation studies were conducted on the two datasets and demonstrate that adopting the calendar of days to be forecasted contributes to the forecasting performance of the Tsformer on tourism demand forecasting. Finally, visualization of the attention weight matrix was performed to reveal the critical information that the Tsformer concerns on the Jiuzhaigou valley dataset. It indicates that the proposed Tsformer can extract dependencies of the tourism demand sequence efficiently, enhancing the interpretability of the proposed Tsfromer.

Arrangements of other parts of this study are as follows. Section 2 introduces the related work of this study. Section 3 describes valina Transformer and the proposed Tsformer. In section 4, we present experiments among the proposed Tsformer and nine baseline models on the Jiuzhaigou valley and Siguniang mountain datasets, as well as corresponding ablation studies of the Tsformer are performed on the two datasets. Section 5 analyzes the interpretability of the Tsformer. Section 6 discusses the conclusion of this study and the future work.

## 2. Related work

*2.1. Tourism demand forecasting models*

Over 600 studies on tourism demand forecasting have been published in the past decades; some studies have been adopted in the industry, which benefits governments and enterprises [15]. The current forecasting methods can be divided into four classes: subjective approaches, time series models, econometric models, and AI-based models. Some studies used multiple methods or a combination of them to enhance forecasting performance.

Time series models can be divided into basic time series models and advanced time series models [16]. Basic time series models include Naïve, Auto Regression (AR), Moving Average (MA), Exponential Smoothing (ES), Historical Average (HA), etc. Generally, the Naïve method can get better performance in the dataset that does not change drastically in short-term patterns [17]. However, in the drastic changing dataset or long-term forecasting, the performance of Naïve method will decline dramatically [18]. Naïve method is widely adopted by studies as a primary benchmark for evaluating the proposed methods. Basic time series models are unable to harness seasonal features and require the input sequence to be stationary. Nevertheless, tourism activities have strong seasonal features, and seasonal features are considered to be an essential feature of tourism demand forecasting [3]. Compare to basic time series models. Advanced time series models harness the trend and seasonal features. The advanced time series models include seasonal Naive、ARIMA、ARIMAX、SARIMA、SARIMAX, etc. Seasonal Naïve uses the historic ground truth as prediction [19], SARIMA is a variant of ARIMA with seasonality, and ARIMAX is ARIMA with exogenous variable. Based on ARIMAX and SARIMA, SARIMAX harnesses seasonal features and exogenous variables simultaneously, which have become one of the most widely used models in tourism demand forecasting.

Econometric models can be divided into static econometric models and dynamic econometric models [16]. Static econometric models include linear regression [20], gravity model [21], etc. Vector autoregressive (VAR) [22], the error correction models (ECM) [23] and time-varying parameter (TVP) [24] are the three representative models of dynamic econometric models. Compared to static econometric models, dynamic econometric models can capture the time-varying of consumer preferences and enhances the forecasting performance of econometric models.

When processing a large amount of data, generally, AI-based methods can obtain better performance than traditional methods. The excellent performance of AI-based methods could rely on its internal feature engineering ability (Law, Li, Fong, & Han, 2019). However, AI-based methods are considered black boxes, i.e., AI-based methods are weak in interpretability. In the past years, with the development of AI, more AI-based methods started to be adopted by studies, e.g., Support Vector Regression (SVR), k-nearest neighbor (k-NN), Artificial Neural Network (ANN) and Recurrent Neural Network (RNN). Compared to linear regression, hidden layers and activation function of ANN bring excellent non-linear function fitting ability [8, 9]. Nevertheless, the shortcoming of ANN for time series forecasting is that ANN does not have the good architecture to process time series [9]. Therefore, RNN started to be adopted by some studies. Long Short-Term Memory (LSTM) is an improved variant of RNN. Compared to RNN, LSTM adds cell state to memorize long-term dependency, and uses input gate, output gate, and forget gate to handle cell state and mitigate gradient vanishing [25]. LSTM is currently adopted in some tourism demand forecasting studies and obtained high forecasting performance. Lv, Peng, & Wang use LSTM to forecast the tourism demand of America, Hainan, Beijing, and Jiuzhaigou valley [26]. In the study by B. Zhang, Pu, Wang, & Li, LSTM is adopted in hotel accommodation forecasting [27]. Except for LSTM, some variants of LSTM have been used in related studies. Bi-directional LSTM(Bi-LSTM) is considered to perform better in processing long-term dependency. In a recent study by Kulshrestha, Krishnaswamy, & Sharma, Bi-LSTM is used for forecasting tourism demand in Singapore [28].

*2.2. Search engine index*

In 2012, hotel room demand forecasting [29] and travel destination planning forecasting [30] adopted the Google trend and proved that the adoption of the search engine index enhances the forecasting performance. Subsequent studies adopt the Google trend and the Baidu index. The Baidu index is used more in tourism activities related to China, while the Google trend is used more in tourism activities of Europe and America [4]. According to the study by Yang, Pan, Evans, & Lv, the Baidu index in China-related studies performs better than the Google trend [31]. In recent studies, the number of keywords on the search engine index has increased. X. Li et al. uses 45 keywords to forecast tourism demand in Beijing; the Pearson correlation coefficients between these indexes and tourism demand are calculated, which indicates that tourism demand does not strongly correlate with all indexes. To avoid loss of features, all search engine indexes are preserved and principal component analysis (PCA) is adopted to perform dimensionality reduction [7]. In the study by Law, Li, Fong, & Han, up to 256 keywords are used for Macau tourist arrival volumes forecasting [32]. Researchers hope to represent every aspect of tourists' interests in tourism destinations by covering multi keywords related to tourism destinations.

*2.3. Transformer*

In 2017, Vaswani et al. proposed Transformer to solve machine translation [33], then Transformer is used for constructing large-scale language models such as BERT. BERT achieves state-of-the-art in multiple tasks [34]. Benefit from the excellent performance on long sequence, Transformer is introduced into computer vision. Related studies have been conducted on image classification [35], object detection [36], object tracking [37], and semantic segmentation [38]. Recently, few studies try to exploit the potential of Transformer for time series forecasting. S. Li et al. enhances the locality and reduces the memory usage of the Transformer in time series forecasting [39]. Lim, Arik, Loeff, & Pfister combines LSTM and Multi-Head attention for time series forecasting [40]. Although Transformer is proved to outperform LSTM in multi fields, there is no study that introduces Transformer into tourism demand forecasting yet.

## 3. Methodlogy

*3.1. Transformer*

Fig. 1 presents the basic architecture of Transformer; Transformer is an Encoder-Decoder model based on multi-head attention. The encoder generates the corresponding syntax vector of the input sequence, while the decoder generates the target sequence according to both the input sequence of the decoder and the output syntax vector of the encoder. In recent studies, generally, LSTM or GRU will be adopted in sequence modeling. They are RNN essentially, while Transformer is a network that differs from RNN; the core of Transformer is the multi-head attention mechanism, which allows computing attention of multiple time steps simultaneously and does not require state accumulation in order.

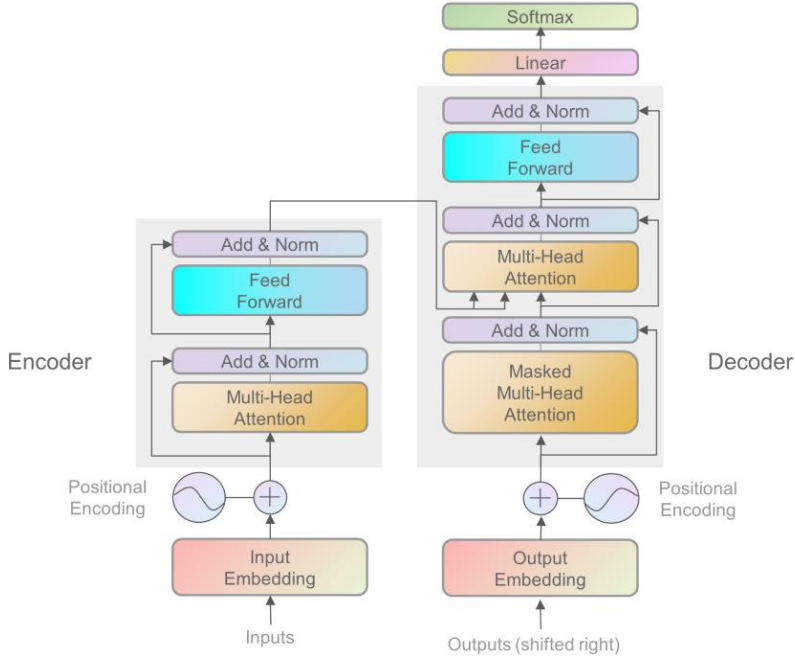

Fig. 1. The architecture of valina Transformer

### 3.1.1. Multi-head attention

Generalized attention can be expressed as:

$$\text{attention output} = \text{Attention}(Q, K, V) \tag{1}$$

In Eq. 1, Q, K, V represent query, key and value in attention mechanism, the weight of value is calculated through query and key, then the calculated weight is used to perform a weighted sum of value. The attention calculation in Transformer is implemented by scaled dot-product attention, which is shown in Fig. 2(a). Scaled dot-product attention can be denoted as:

$$\text{attention}(Q, K, V) = softmax\left(\frac{QK^T}{\sqrt{d_k}}\right)V \tag{2}$$

In Eq. 2, $d_k$ represents the dimension of key

Fig. 2(b) presents the calculation of multi-head attention; in the multi-head attention mechanism, the calculation of the i-th attention head can be represented as:

$$head_i = \text{attention}(QW_i^Q, KW_i^K, VW_i^V) \tag{3}$$

In Eq. 3, $W_i^Q$, $W_i^K$ and $W_i^V$ denote the linear transformation of Q, K, V of the i-th attention head.

Multi-head attention is the concatenation of each attention head. Let n denote the number of attention heads; multi-head attention can be denoted as:

$$\text{MultiHead}(Q, K, V) = \text{concat}(head_1, \ldots, head_n)W^O \tag{4}$$

In Eq. 4, concat represents to concatenate operation. $W^O$ represents the linear transformation of concatenated output.

The internal multi-head attention has helped Transformer to overcome the shortcoming of non-parallel training and the bottleneck of the memory capacity in processing long-term dependency of LSTM or GRU.

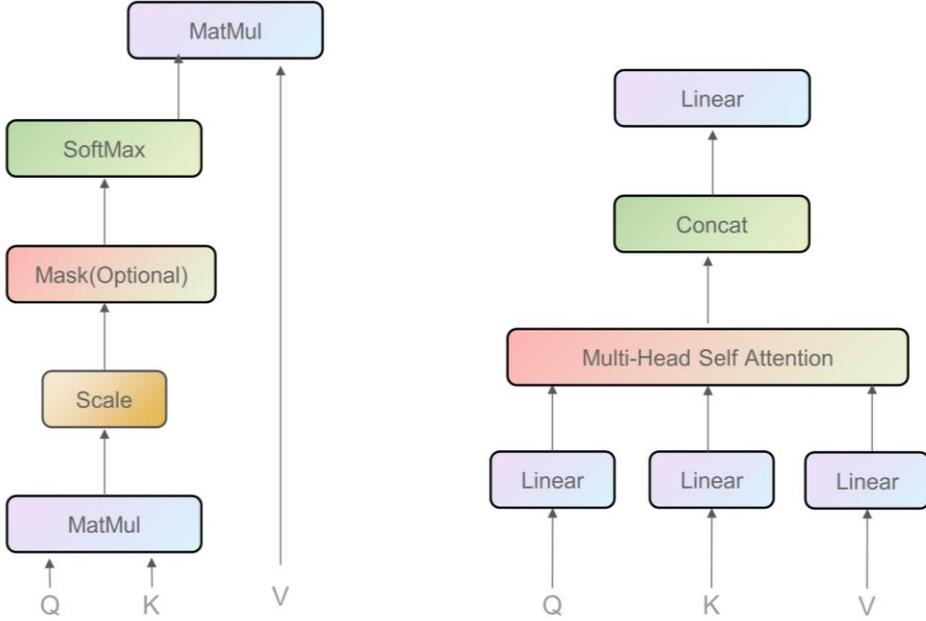

Fig. 2. (a) Scaled Dot-Product Attention; (b) Multi-Head Self Attention

*3.1.2. Feed-forward network*

The feed-forward network (FFN) in Fig. 1 consists of two linear transformations and Rectified Linear Unit (ReLU) activation function. The computation of the feed-forward network is position-wise, i.e., the weight of linear transformation of each time step is identical. The formula of the feed-forward network can be given as:

$$\text{FFN}(x) = \max(0, xW_1 + b_1)W_2 + b_2 \tag{5}$$

ReLU activation function is used for mitigating gradient vanishing, gradient explosion and accelerating convergence. However, all negative values of output will be restricted to zero, which may cause the problem of neuron inactivation, and no valid gradient will be obtained in backward propagation to train the network. Consequently, we replaced ReLU activation function with Exponential Linear Unit(ELU) activation function [41]. Differs from ReLU, ELU has negative output, which ensures ELU is more robust to noise. ELU and feed-forward network with ELU can be denoted as:

$$\text{ELU}(x) = \begin{cases} \alpha(e^x - 1), & x \leq 0 \\ x, & x > 0 \end{cases} \tag{6}$$

$$\text{FFN}(x) = ELU(xW_1 + b_1)W_2 + b_2 \tag{7}$$

*3.1.3. Masked multi-head attention*

Masked multi-head attention includes an optional masking mechanism, which determines the way of attention calculation, i.e., prevents attention from attending time steps in specific locations. In general, the mask in masked multi-head attention is an m × n matrix. In self-attention, m equals n, and the mask is a square matrix; m or n denotes the length of the input sequence. Moreover, in encoder-decoder attention, m denotes the length of decoder self-attention results, n denotes the length of encoder output. The formula of

scaled dot-product attention in masked multi-head attention can be given as:

$$\text{Attention}(Q, K, V) = softmax\left(\frac{QK^T + M}{\sqrt{d_k}}\right) V \quad (8)$$

In Eq. 8, M denotes mask matrix, 0 elements in the mask represent the corresponding time steps that will be attended, while negative infinity elements represent the corresponding time steps that will be ignored in attention calculation. After softmax in scaled dot-product attention, the corresponding attention weight of each negative infinity element will be infinitely close to 0, so any information from the masked time steps will not flow into the current time step.

### 3.2. Tsformer

#### 3.2.1. Input and output of Tsformer

For model input, rolling window strategy is adopted, that is, take the data for a period of time iteratively with a step size of 1 according to specific window size. Three parameters are used to control the rolling window size and the number of steps to be forecasted, which are encoder input length, decoder input length, and forecast horizon. When performing 1-day ahead forecasting, the three parameters are set to 7, 5 and 1, respectively. Fig. 3 shows the input and output of the Tsformer; $T_i$ denotes the i-th day; $P_i$ denotes the forecasting of the i-th day. The encoder and decoder input of the Tsformer in Fig. 3 contains some overlap, aiming to reuse the potential raw features of short-term dependency to prevent degrading of attention to short-term dependency.

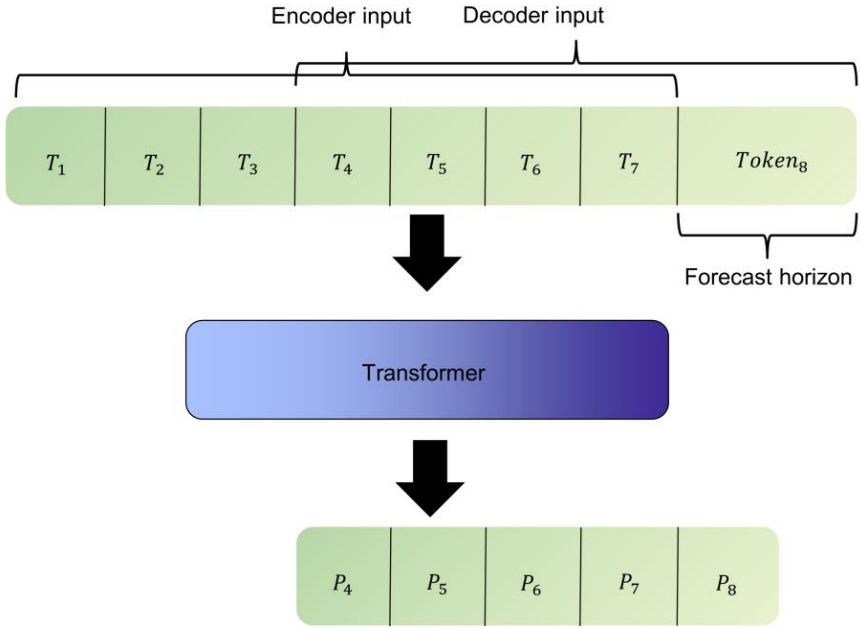

Fig. 3. Input and output of the Tsformer

Differs from other models, models with Encoder-Decoder architecture are used for the sequence-to-sequence task; the output of the proposed Tsformer is a sequence whose length is determined by decoder input length.

In the prediction stage, the decoder obtains information from the previous time steps to generate the next time step. In order to generate the decoder output $P_8$, $Token_8$ which plays the role of a placeholder is required to represent the unknown $T_8$, the number of tokens is determined by the forecast horizon, and scalar 0 is used to fill the tokens.

### 3.2.2. Input layer

Input layers of the encoder and decoder in valina Transformer are implemented by embedding layer, which transforms the discrete variables into dense vectors. With the training progress, the corresponding dense vectors of discrete variables will present a particular distribution in the vector space; their distance in vector space can measure the difference of them.

Valina Transformer has an excellent performance in machine translation; it accepts vocabulary ID as input. As the frequency of each ID in vocabulary is high in the corpus, valina Transformer has a promising ability to distinguish dense vectors after embedding mapping. However, in time series forecasting, most of the variables in the input sequence are continuous, and a regression problem is required to be solved. However, the embedding layer can only transform discrete variables into dense vectors. In the proposed Tsformer, the embedding layer in valina Transformer is replaced by a fully connected layer.

### 3.2.3. Positional Encoding

Hidden states of LSTM or GRU are calculated through the accumulation of states; LSTM or GRU can naturally represent the location information of time steps. In the attention mechanism of Transformer, the attentions of all time steps are calculated simultaneously. If the time steps in Transformer are not distinguished, the attention calculation at all time steps will get the same result. In general, additional information of position is required by Transformer. The additional information of position is introduced by positional encoding after the input layer.

There are two commonly used positional encodings, positional embedding in BERT and sinusoid positional encoding in valina Transformer; the formula of sinusoid positional encoding can be denoted as:

$$PE_{(pos,2i)} = \sin\left(\frac{pos}{10000^{2i/d_{model}}}\right) \tag{9}$$

$$PE_{(pos,2i+1)} = \cos\left(\frac{pos}{10000^{2i/d_{model}}}\right) \tag{10}$$

$d_{model}$ denotes the dense vector dimension of each time step mapped by the input layer; this dimension is consistent with the dimension of input and output of the encoder and decoder. 2i and 2i + 1 represent the even and odd dimensions in the $d_{model}$, respectively. $pos$ represents the position index of the input time step. Let the length of the input sequence be N, then the range of pos is $[0, N-1]$.

The purpose of positional embedding is to implicitly learn a richer position relationship. However, the datasets in tourism demand forecasting are generally small; it will be harder for the learned positional embedding to learn the relative relationship of time steps. Compared to positional embedding, the advantage of sinusoid positional encoding is that the encoding of position is fixed for each time step, and training of additional weights is not required. Therefore, sinusoid positional encoding is adopted in the proposed Tsformer to encode the position of time steps explicitly. After positional encoding is generated, it is added with the output of the input layer position-wisely, as shown in Fig 4.

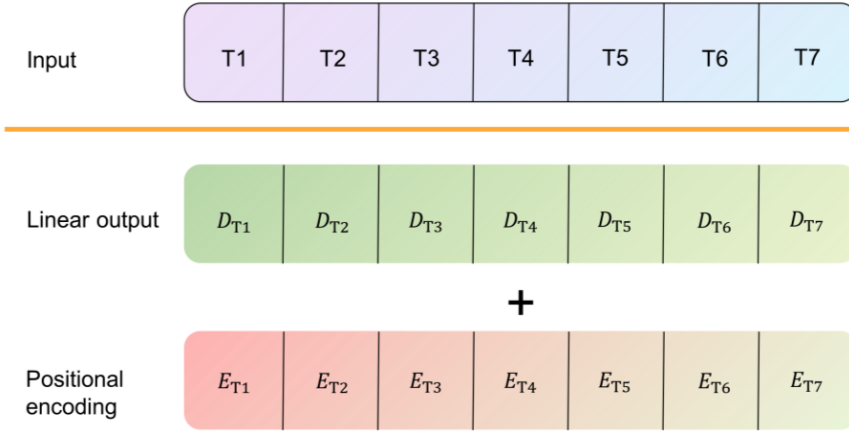

Fig. 4. Positional encoding on the output of the input layer.

*3.2.4. Attention masking mechanism*

Context matters in text sequence; in valina Transformer, attentions of each time step will be calculated with other time steps to capture context information, which leads to complex attention interactions. However, time steps in time series are only related to the previous time steps of them. Global attention of every time step may result in introducing redundant information and prevents from discovering the dominant attentions in attention visualization. Therefore, encoder source masking, decoder target masking, and decoder memory masking are adopted to simplify attention interactions in time series processing to avoid redundant information and highlight the dominant attentions.

**Encoder source masking**

In the calculation of self-attention in the encoding stage, the encoder source masking is adopted to ensure that every time step only includes previous information of themselves to simplify the attention interactions. Not like LSTM, the hidden state of each time step does not rely on state accumulation, which is calculated in parallel. If long-term dependencies exist, since the hidden states of earlier time steps are restricted to contain less information, the attention to earlier time steps in encoder output will be more explicit in the encoder-decoder attention. Encoder source mask in 1-day ahead forecasting can be given as:

$$Mask = \begin{pmatrix} 0 & -inf & -inf & -inf & -inf & -inf & -inf \\ 0 & 0 & -inf & -inf & -inf & -inf & -inf \\ 0 & 0 & 0 & -inf & -inf & -inf & -inf \\ 0 & 0 & 0 & 0 & -inf & -inf & -inf \\ 0 & 0 & 0 & 0 & 0 & -inf & -inf \\ 0 & 0 & 0 & 0 & 0 & 0 & -inf \\ 0 & 0 & 0 & 0 & 0 & 0 & 0 \end{pmatrix} \quad (11)$$

In Eq. 11, $-inf$ represents negative infinity

Take the calculation of the fifth time step of encoder self-attention in 1-day ahead forecasting as an example; the green arrows represent the time steps pointed participate in the attention calculation, while the red arrows represent the time steps pointed are not attended. Let $Enc_{Ii}$ denote the i-th input time step of encoder input. Fig. 5 shows in the calculation of $Enc_{I5}$, the time steps from $Enc_{I1}$ to $Enc_{I5}$ are attended.

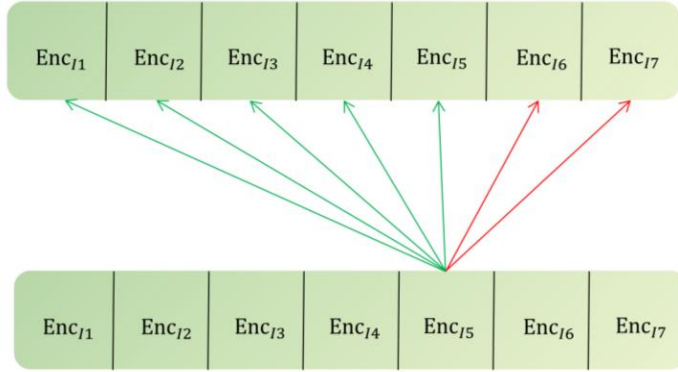

Fig. 5. Encoder source masking

**Decoder target masking**

In the calculation of self-attention in the decoder, it is also necessary to prevent attention from attending the follow-up time steps of the current time step. In multi-step ahead forecasting, multiple tokens are included in the decoder input. When decoding the current token, the follow-up tokens of the current token are equivalent to padding in the NLP models, which do not contribute to the forecasting of the current token and introduce redundant information. Therefore, decoder target masking is adopted to exclude the follow-up tokens of the current token properly. Decoder target mask in 1-day ahead forecasting can be represented as:

$$Mask = \begin{pmatrix} 0 & -inf & -inf & -inf & -inf \\ 0 & 0 & -inf & -inf & -inf \\ 0 & 0 & 0 & -inf & -inf \\ 0 & 0 & 0 & 0 & -inf \\ 0 & 0 & 0 & 0 & 0 \end{pmatrix} \quad (12)$$

Let $Dec_{Ii}$ denote the i-th time step of the decoder input. Fig. 6 shows that the actually attended time steps in decoder self-attention in the calculation of $Dec_{I5}$ in 1-day ahead forecasting are $Dec_{I4}$ and $Dec_{I5}$.

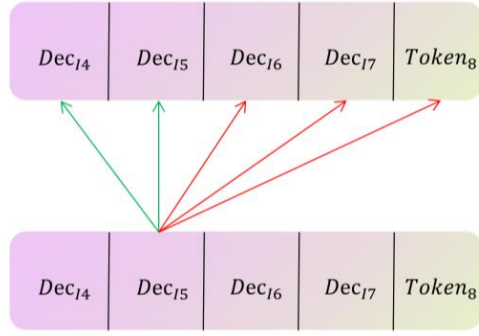

Fig. 6. Decoder target masking

**Decoder memory masking**

Since there is some overlap in the encoder input and decoder input, when calculating the encoder-decoder attention between calculation results of the decoder self-attention and encoder output, it is necessary to prevent the unattended time steps in the encoder to be attended in the decoder through encoder-decoder attention. Therefore, decoder memory masking is adopted to keep the current time step in calculation results of the decoder from attending the follow-up time steps of the corresponding time step in the encoder output. Decoder memory mask in 1-day ahead forecasting can be represented as:

$$Mask = \begin{pmatrix} 0 & 0 & 0 & -inf & -inf & -inf & -inf \\ 0 & 0 & 0 & 0 & -inf & -inf & -inf \\ 0 & 0 & 0 & 0 & 0 & -inf & -inf \\ 0 & 0 & 0 & 0 & 0 & 0 & -inf \\ 0 & 0 & 0 & 0 & 0 & 0 & 0 \end{pmatrix} \quad (13)$$

Let $Enc_{oi}$ denote the i-th time step in the encoder output, and $Dec_{si}$ denotes the i-th time step in calculation results of decoder self-attention. Fig. 7 presents that when calculating encoder-decoder attention between $Dec_{S5}$ and the encoder output in 1-day ahead forecasting, the actually attended time steps in the encoder output are from $Enc_{o1}$ to $Enc_{o4}$.

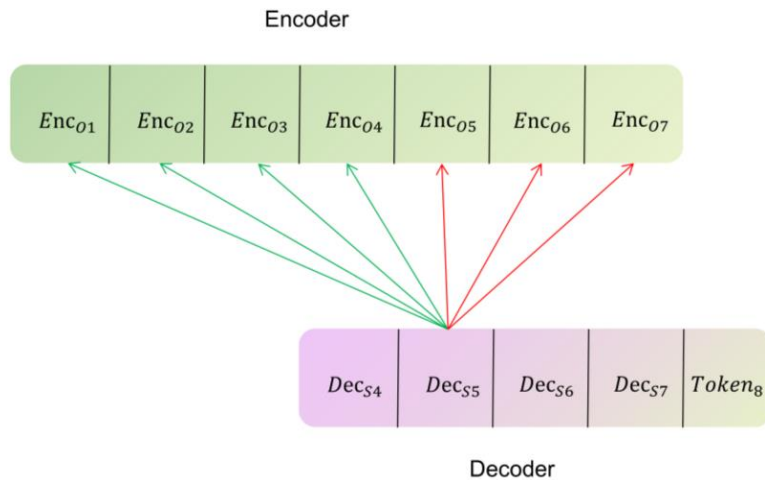

Fig. 7. Decoder memory masking

### 3.2.5. Calendar of days to be forecasted

In Section 3.2.1, zero-filled vectors are used for tokens to represent time steps to be forecasted. However, in tourism demand forecasting, information of time days to be forecasted is not all unknown; the calendar of days to be forecasted can be obtained before forecasting. Weekday and month are adopted to represent the calendar for better learning of seasonality. Therefore, the calendar of days to be forecasted is added to tokens, leaving the unknown tourism demand and search engine index filled by zero to enhance the forecasting performance of the proposed Tsformer model. Fig. 8 presents the adoption of the calendar of $Token_8$ in 1-day ahead forecasting.

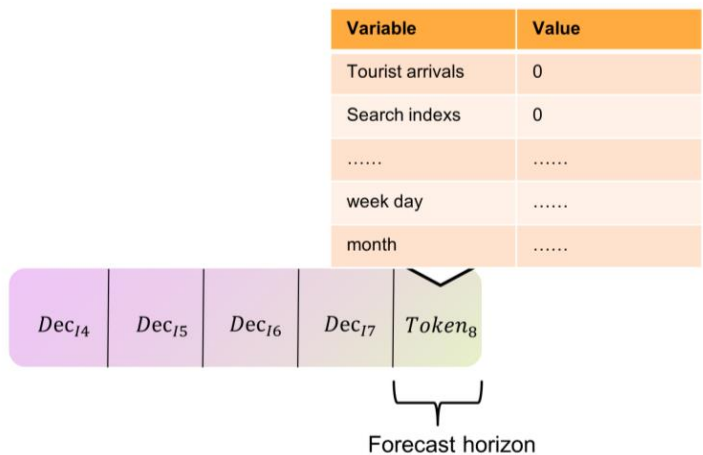

Fig. 8. Adoption of the calendar of days to be forecasted

*3.2.6. Output layer*

The output of valina Transformer is activated by softmax and converted to the confidence of IDs in vocabulary. In this case, valina Transformer solves the classification problem. The regression problem should be solved in tourism demand forecasting, so softmax is removed from the last layer.

The architecture of the proposed Tsformer is shown in Fig. 9:

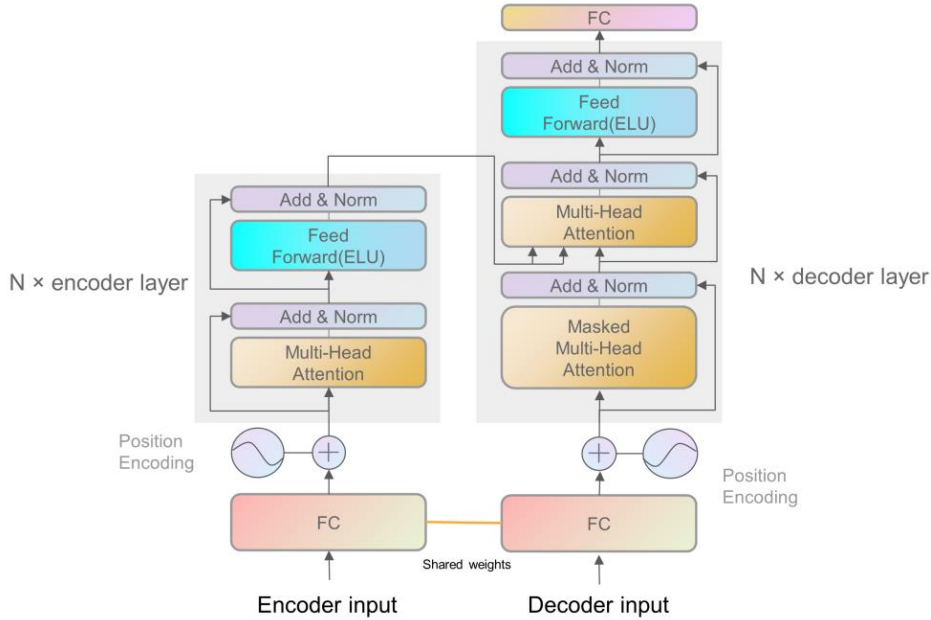

Fig. 9. The architecture of the proposed Tsformer

## 4. Experiments

*4.1. Datasets*

As shown in Fig. 10, tourism demands from January 1, 2013 to June 30, 2017 in Jiuzhaigou valley and September 25, 2015 to December 31, 2018 in Siguniang mountain were collected. As the search engine index is considered the main feature of tourism demand forecasting, we used Jiuzhaigou valley and Siguniang mountain as keywords and collected the corresponding Baidu index, as well as the other 11 related keywords. Table 1 shows all keywords in the two datasets and the Pearson correlation coefficients between their corresponding Baidu indexes and the tourism demand. The tourism demand strongly correlates with the search indexes corresponding to other keywords except for Jiuzhaigou ticket.

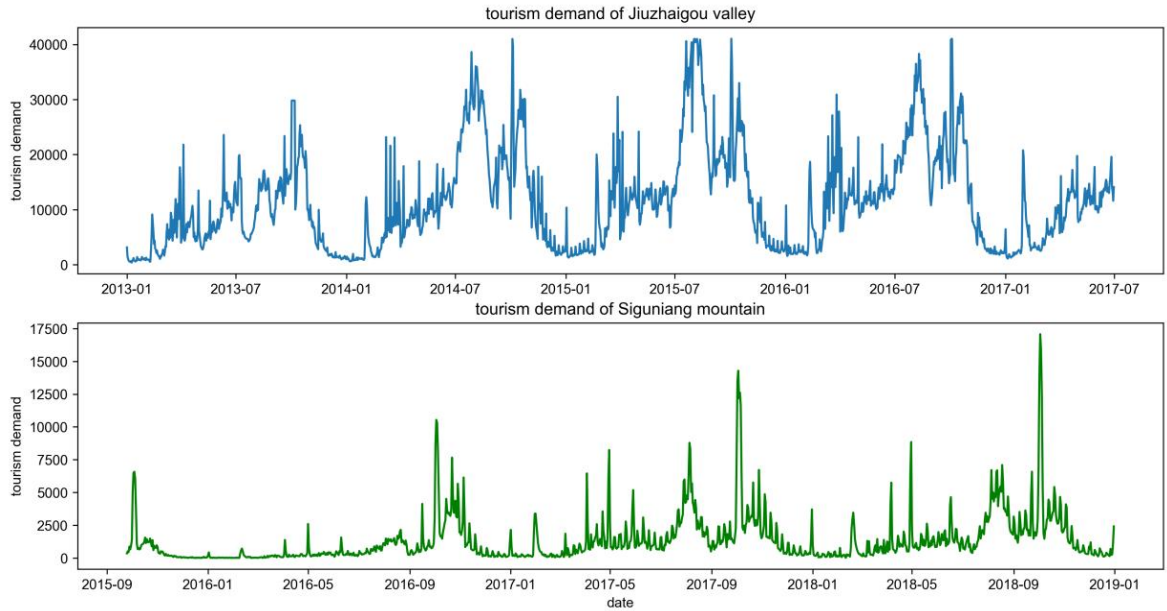

Fig. 10. The tourism demand of Jiuzhaigou valley and Siguniang mountain

Table 1. Keywords and Pearson correlation coefficients

| Dataset | | | |
|---|---|---|---|
| Jiuzhaigou valley | | Siguniang mountain | |
| Keyword | Pearson correlation coefficient | Keyword | Pearson correlation coefficient |
| Jiuzhaigou | 0.6973 | Siguniang | 0.6006 |
| Jiuzhaigou tips | 0.7546 | Siguniang lodging | 0.5409 |
| Jiuzhaigou hotel | 0.2557 | Siguniang photo | 0.4365 |
| Jiuzhaigou brief introduction | 0.3727 | Siguniang weather | 0.6891 |
| Jiuzhaigou introduction | 0.2949 | Siguniang weather forecast | 0.5819 |
| Jiuzhaigou lodging | 0.6894 | Siguniang tips | 0.5350 |
| Jiuzhaigou route | 0.6412 | Siguniang travel | 0.3833 |
| Jiuzhaigou travel tips | 0.8051 | Siguniang travel tips | 0.4849 |
| Jiuzhaigou map | 0.8398 | Siguniang scenic area | 0.4574 |
| Jiuzhaigou weather | 0.6567 | Siguniang elevation | 0.6759 |
| Jiuzhaigou ticket | 0.0330 | Siguniang road trip | 0.3467 |
| Jiuzhaigou travel | 0.6541 | Siguniang ticket | 0.6386 |

For richer features in tourism demand forecasting, daily weather, highest temperature, wind, air quality were collected. Since the annual air quality and wind in Jiuzhaigou valley and Siguniang mountain is stable, they are excluded. Date type, month and weekday were included in the two datasets considering the influence of holiday, seasonality, as well as low-season and high-season of tourism demand.

The two datasets were divided into train set, validate set and test set. On the Jiuzhaigou valley dataset, data from January 1, 2013 to June 30, 2015 were used as the train set; July 1, 2015 to June 30, 2016 were in the validate set; July 1, 2016 to June 30, 2017 were in the test set. For the Siguniang mountain dataset, September 25, 2015 to December 31, 2017 were in the train set; January 1, 2018 to June 30, 2018 were in the validate set; July 1, 2018 to December 31, 2018 were in the test set.

*4.2. Dummy variable encoding*

The tourism demand datasets of Jiuzhaigou valley and Siguniang mountain contain categorical variables such as weather and date type, which can not be used directly by the model. In order to handle these variables properly, these variables should be encoded into dummy variables. Date type can be divided into working day, weekend and holiday, as shown in Table 2.

Table 2. Date type encoding

| Date type   | Dummy variable |
|-------------|----------------|
| Working day | 0              |
| Weekend     | 1              |
| Holiday     | 2              |

There are many categories of weather, and more categories will lead to the risk of overfitting. Similar weather was merged into five categories, as shown in Table 3.

Table 3. Weather Encoding

| Weather  | Value |
|----------|-------|
| Snow     | 0     |
| Rain     | 1     |
| Overcast | 2     |
| Cloudy   | 3     |
| Sunny    | 4     |

Month and weekday were used to represent the calendar, i.e., January to December were encoded as 0 to 11. Moreover, Monday to Sunday were encoded as 0 to 6.

*4.3 Preprocessing*

For the models trained by gradient descent algorithm, the dimensions of some variables in the datasets are inconsistent. In order to avoid over-large variables dominating the output of the models and accelerate the convergence of the models, it is necessary to normalize the data to make the variables comparable. Min-Max normalization was adopted, which can be denoted as:

$$X_{norm} = \frac{X - X_{min}}{X_{max} - X_{min}} \tag{14}$$

## 4.4. Metrics

To compare the performance of models, MAE (mean absolute error), RMSE (root mean squared error), and MAPE (mean absolute percentage error) were used as metrics, which can be given as:

$$\text{MAE} = \frac{1}{n}\sum_{i=1}^{n}|\hat{y}_i - y_i| \tag{15}$$

$$\text{RMSE} = \sqrt{\frac{1}{n}\sum_{i=1}^{n}(\hat{y}_i - y_i)^2} \tag{16}$$

$$\text{MAPE} = \frac{100\%}{n}\sum_{i=1}^{n}\left|\frac{\hat{y}_i - y_i}{y_i}\right| \tag{17}$$

In Eq.15, Eq. 16 and Eq. 17, $\hat{y}_i$ denotes the predicted value; $y_i$ denotes the ground truth.

## 4.5. Model settings

Most of the hyperparameters were obtained through grid search. The hyperparameters of the proposed Tsformer and other baseline models in 1-day ahead forecasting on the Jiuzhaigou valley and Siguniang mountain dataset are given in Table 4 and Table 5; the hyperparameters in other forecast horizons are similar to them. For better performance of baseline models, some adjustments were made.

**(seasonal) Naïve**

In 1-day ahead forecasting, Naïve method uses the ground truth at time $t-1$ as the forecast at time $t$. However, the standard Naïve method is not suitable for multi-day ahead forecasting. Therefore, in multi-day ahead forecasting, seasonal Naïve was used to replace the standard Naïve, which uses the ground truth at time $t-s$ as the forecast at time t. If time $t-s$ is still in the forecast horizon, then the forecasts are given according to the valid ground truth from earlier periods.

**ARIMA series models**

According to BIC (Bayesian Information Criterion), the hyperparameters of ARIMA series models were selected on the train set. ACF and PACF were used to determine the approximate range of the hyperparameters, and a grid search is performed to select the model with the lowest BIC. To handle multi-day ahead forecasting, out-of-sample forecasts were used in place of lagged dependent variables. For relative fairness in comparison with other models, rolling-origin evaluations [42] were also applied to multi-day ahead forecasting of ARIMA series models.

**SVR**

SVR only supports single-target regression, and it is unable to handle multi-day ahead forecasting natively. In order to perform multi-day ahead forecasting with SVR, multi SVR models were trained to handle each day to be forecasted.

Table 4. The hyperparameters of the Tsformer

| Hyperparameter | Dataset | |
|---|---|---|
| | Jiuzhaigou valley | Siguniang mountain |
| Encoder input length | 7 | 7 |
| Decoder input length | 5 | 5 |
| $d_{model}$ | 32 | 32 |
| Heads | 4 | 4 |
| Learning rate | 3e-3 | 5e-3 |
| Encoder layers | 4 | 1 |
| Decoder layers | 4 | 1 |
| Feed-forward dimension | 64 | 128 |
| Dropout | 0.1 | 0.1 |

Table 5. The hyperparameters of the baseline models

| Model | Dataset | |
|---|---|---|
| | Jiuzhaigou valley | Siguniang mountain |
| (seasonal) Naïve | S=1 for single step, S=7 for multi-step | |
| ARIMA | p=4, d=1, q=4 | p=4, d=1, q=3 |
| ARIMAX | p=4, d=1, q=4 | p=4, d=1, q=3 |
| SARIMA | p=4, d=1, q=4, P=1, D=0, Q=1, S=7 | p=4, d=1, q=3, P=0, D=1, Q=1, S=7 |
| SARIMAX | p=4, d=1, q=4; P=1, D=0, Q=1, S=7 | p=4, d=1, q=3, P=0, D=1, Q=1, S=7 |
| SVR | kernel=rbf, gamma=0.01, C=1, epsilon=1e-5 | kernel=rbf, gamma=0.001, C=10, epsilon=1e-5 |
| k-NN | neighbors=5, leaf size=5, p=2 | neighbors=10, leaf size=5, p=1 |
| ANN | hidden layers=2, learning rare=1e-2, hidden size=128 | |
| LSTM | hidden size=128, learning rare=1e-2 | |

## 4.6. Experiments results

Table 6 presents the results of the experiments of the proposed Tsformer and other nine baseline models on the Jiuzhaigou valley and Siguniang mountain datasets; forecast horizons are set to 1, 7, 15 and 30.

### 4.6.1. The Jiuzhaigou valley dataset

As shown in Table 6, when performing 1-day ahead forecasting on the Jiuzhaigou valley dataset, Naïve method has excellent performance. Although ARIMA series models have not achieved the performance that exceeds Naïve method, they outperformed SVR, k-NN and ANN, and performed better than LSTM under MAE and MAPE, which indicates that ARIMA series models have more advantages compared to other machine learning models in 1-day ahead forecasting. LSTM gets the lowest RMSE, but the Tsformer gets the lowest MAE and MAPE at the same time. The Tsformer only fails to outperform Naïve method and LSTM under RMSE. For 7-day ahead forecasting, seasonal Naïve no longer gets high performance; it only outperforms k-NN under MAPE. ARIMA series models still get promising performance, i.e., they outperform k-NN, and their performance is close to that of ANN. However, ARIMA series models get higher MAE and RMSE than SVR and fall behind LSTM in all metrics. The Tsformer is slightly behind SVR under RMSE but outperforms other models under MAE and MAPE. When performing 15-day ahead forecasting, ANN, LSTM and the Tsformer are the only three models that perform well. Overall, the performance of LSTM is better than that of ANN, and the Tsformer shows significant advantages over LSTM in performance. The proposed Tsformer outperforms all other baseline models under all metrics. The results of 30-day ahead forecasting indicate that models except for LSTM and the Tsformer do not have excellent long-term forecasting ability. The proposed Tsformer outperforms LSTM under MAE and RMSE and significantly outperforms LSTM under MAPE, achieving the best performance compared to all baseline models.

Fig. 11 presents the performance of each model in the experiments on the Jiuzhaigou valley dataset under MAE, RMSE and MAPE. Naïve method only gets excellent short-term forecasting performance but faces a rapid decline in performance in long-term forecasting. ARIMA, ARIMAX, SARIMA and SARIMAX get similar performance. In short-term forecasting, ARIMA series models outperform all other models except for the Naïve method and the proposed Tsformer but face a rapid decline in performance in the long-term forecasting of 15 days or more. ARIMA series models underperform in long-term forecasting. K-NN is outperformed by other models in 1-day to 7-day ahead forecasting, although it gets low MAE and RMSE in 15-day to 30-day ahead forecasting, it is outperformed by ANN, LSTM and the Tsformer. The performance of SVR is close to ARIMA series models in 1-day to 7-day ahead forecasting but is outperformed by ANN, LSTM and the Tsformer in other forecast horizons. Although LSTM has no significant advantages in short-term forecasting, it still outperforms ANN. In long-term forecasting, the structural advantage of LSTM as a recurrent neural network has been reflected. The overall error curves rise slowly under the three metrics, but the best performance is still not achieved by LSTM. For forecast horizons from 1 to 30, the error curves of the proposed Tsformer rise more slowly, and they are almost lower than that of other baseline models. On the Jiuzhaigou valley dataset, ARIMA series models, LSTM and the Tsformer perform well in short-term forecasting, while only LSTM and the Tsformer perform well in long-term forecasting, and the Tsformer outperforms LSTM. The proposed Tsformer achieves the best performance in both short-term and long-term forecasting tasks on the Jiuzhaigou valley dataset.

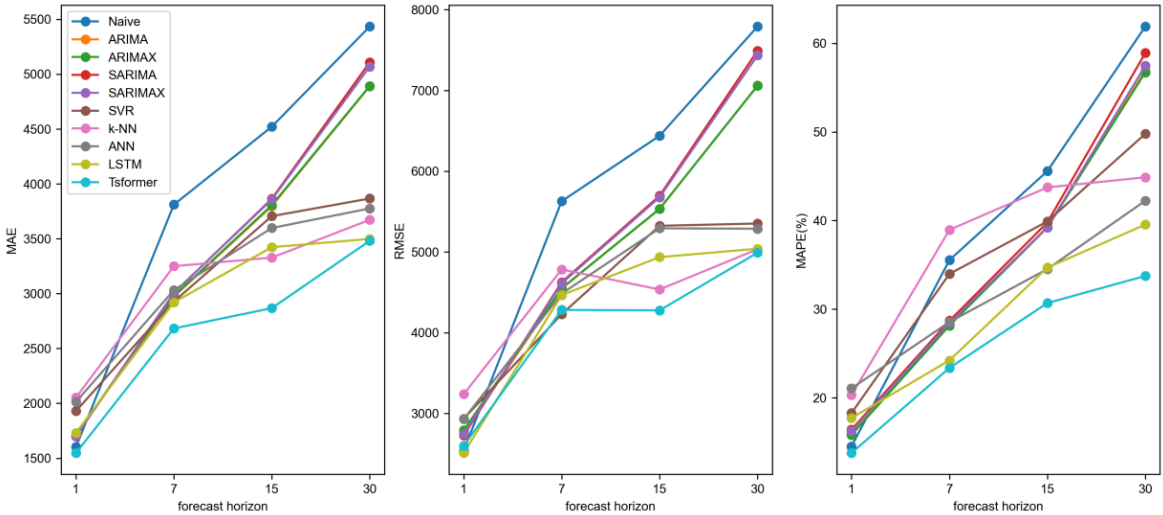

Fig. 11. Performance on Jiuzhaigou valley dataset

*4.6.2. The Siguniang mountain dataset*

As shown in Table 6, when performing 1-day ahead forecasting on the Siguniang mountain dataset, the performance of Naïve method is poor, which is caused by the drastic changes in the tourism demand of Siguniang mountain in short-term. ARIMA series models outperform Naïve method under MAE and RMSE but get higher MAPE than Naïve. The forecasting performance of k-NN is the poorest, and SVR, ANN, LSTM and the Tsformer perform well. Among them, The Tsformer outperforms all baseline models with MAE, RMSE and MAPE of 470.54, 779.67 and 25.51. For 7-day ahead forecasting, the performance degradation of ARIMA series models and k-NN is severe. SVR, ANN, LSTM and the Tsformer still perform well, and SVR outperforms ANN. The Tsformer only falls behind LSTM under RMSE but outperforms LSTM under MAE and MAPE. In 15-day ahead forecasting, there remains ANN, LSTM and the Tsformer achieving relative good performance, among which the Tsformer gets the best performance under MAE, RMSE and MAPE. Although SVR performs well under MAPE, it gets poor performance under MAE and RMSE. In the evaluation of 30-day ahead forecasting, LSTM and the Tsformer are the only two models that perform well. Alouthgh k-NN performs better than LSTM under MAE and RMSE, LSTM significantly outperforms it under MAPE. Although SVR performs better than LSTM under MAPE, its performance under MAE and RMSE is poor. The Tsformer gets MAE, RMSE and MAPE of 1291.55, 2276.43 and 64.88, which dramatically surpasses LSTM and achieves the best performance.

Fig. 12 presents the performance of each model in the experiments on the Siguniang mountain dataset under MAE, RMSE and MAPE. Naive method and seasonal Naïve method get poor performance in both short-term and long-term forecasting. MAE and RMSE of Naïve method in 1-day ahead forecasting is only lower than that of k-NN. MAE and RMSE of seasonal Naive method are always higher than that of other models when the forecast horizons are 7 to 30. Naive method and seasonal Naïve method only outperform ARIMA series models under MAPE. Similar to the experiments on the Jiuzhaigou dataset, ARIMA, ARIMAX, SARIMA and SARIMAX still achieve similar performance, but the difference is more prominent. ARIMAX performs better under MAE and RMSE, while SARIMA and SARIMAX perform better under MAPE. However, ARIMA series models do not perform well in long-term or short-term forecasting on the Siguniang mountain dataset compared to other models. K-NN gets poor performance in forecast horizons 1 to

7. In 15-day ahead forecasting, although k-NN is outperformed by ANN, LSTM and the proposed Tsformer under MAE and RMSE, it outperforms ANN under MAPE. When performing 30-day ahead forecasting, k-NN even outperforms LSTM under MAE and RMSE but gets poor performance under MAPE. SVR performs well in 1-day and 7-day ahead forecasting. For 15-day to 30-day ahead forecasting, MAPE of SVR increases slowly, but MAE and RMSE change drastically. The proposed Tsformer only fails to outperform LSTM in 7-day ahead forecasting under RMSE but outperforms all other baseline models. On the Siguniang mountain dataset, SVR, ANN, LSTM and the Tsformer get outstanding performance in short-term forecasting. LSTM and the proposed Tsformer are the only two models that perform well in long-term forecasting, and the Tsformer gets better performance than LSTM in both short-term and long-term forecasting.

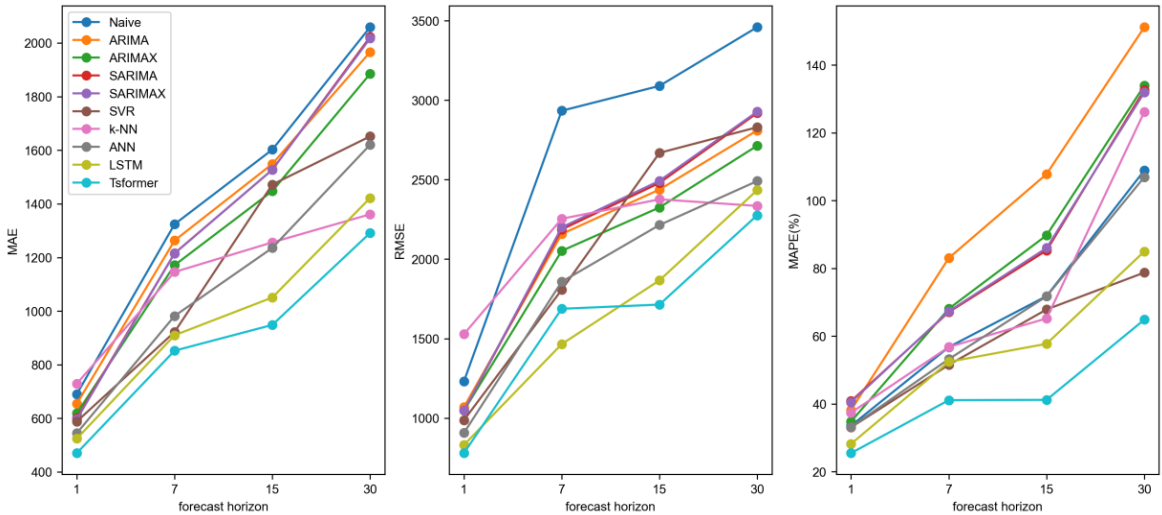

Fig. 12. Performance on the Siguniang mountain dataset

Compared to other models, LSTM and the Tsformer get excellent performance in tourism demand forecasting. With the increasing forecast horizon and input sequence length, information from earlier time steps will vanish in the cell state of LSTM, which leads to decreased forecasting performance. Because of the multi-head attention, the Tsformer does not require accumulating calculations step by step in time series processing but calculates the attentions of each time step simultaneously. Increasing forecast horizon and input sequence length make the performance advantage of the Tsformer in handling long-term dependency more significant.

Table 6. Experiments results on the Jiuzhaigou valley and Siguniang mountain datasets

| Model | Metric | Dataset | | | | | | | |
|---|---|---|---|---|---|---|---|---|---|
| | | Forecast horizon | | | | | | | |
| | | Jiuzhaigou valley | | | | Siguniang mountain | | | |
| | | 1 | 7 | 15 | 30 | 1 | 7 | 15 | 30 |
| (seasonal) Naïve | MAE | 1602.22 | 3810.77 | 4521.16 | 5433.89 | 690.65 | 1324.22 | 1602.41 | 2059.2 |
| | RMSE | 2542.8 | 5628.33 | 6438.76 | 7791.37 | 1231.18 | 2933.76 | 3090.26 | 3459.84 |
| | MAPE (%) | 14.52 | 35.54 | 45.58 | 61.89 | 33.59 | 56.79 | 71.84 | 108.92 |
| ARIMA | MAE | 1706.8 | 2971.98 | 3797 | 4889.13 | 654.71 | 1263.43 | 1549.01 | 1966.29 |
| | RMSE | 2764.85 | 4546.83 | 5531.36 | 7061.46 | 1068.94 | 2158.32 | 2437.89 | 2810.27 |
| | MAPE (%) | 15.94 | 28.27 | 39.29 | 56.89 | 38.24 | 83.07 | 107.84 | 151.18 |
| ARIMAX | MAE | 1715.81 | 2978.82 | 3805.21 | 4892.79 | 619.01 | 1171.41 | 1448 | 1884.98 |
| | RMSE | 2793.75 | 4550.31 | 5534.75 | 7057.23 | 1047.27 | 2052.07 | 2325.3 | 2713.91 |
| | MAPE (%) | 15.81 | 28.15 | 39.21 | 56.72 | 34.83 | 68.11 | 89.76 | 133.96 |
| SARIMA | MAE | 1699.83 | 3005.92 | 3863.69 | 5107.65 | 600.67 | 1215.89 | 1527.38 | 2022.75 |
| | RMSE | 2728.21 | 4625.65 | 5695.82 | 7490.39 | 1047.6 | 2186.75 | 2479.6 | 2919.75 |
| | MAPE (%) | 16.42 | 28.69 | 39.79 | 58.9 | 40.85 | 67.12 | 85.3 | 132.66 |
| SARIMAX | MAE | 1701.51 | 3007.67 | 3857.46 | 5068.46 | 596.45 | 1215.11 | 1527.56 | 2017.22 |
| | RMSE | 2745.95 | 4612.18 | 5677.71 | 7433.91 | 1047.09 | 2200.13 | 2493.15 | 2928.14 |
| | MAPE (%) | 16.24 | 28.45 | 39.22 | 57.47 | 40.53 | 67.34 | 86.02 | 131.92 |
| SVR | MAE | 1929.77 | 2927.45 | 3706.15 | 3866.58 | 588.47 | 922.24 | 1472.18 | 1651.83 |
| | RMSE | 2935.93 | **4229.27** | 5324.13 | 5352.19 | 987.03 | 1807.15 | 2668.84 | 2830.09 |
| | MAPE (%) | 18.28 | 33.98 | 39.9 | 49.78 | 33.13 | 51.56 | 67.89 | 78.77 |
| k-NN | MAE | 2052.4 | 3250.24 | 3327.93 | 3672.22 | 729.61 | 1145.98 | 1256.46 | 1361.28 |
| | RMSE | 3239.64 | 4783.44 | 4535.65 | 5034.09 | 1527.88 | 2254 | 2376.43 | 2334.86 |
| | MAPE (%) | 20.29 | 38.95 | 43.74 | 44.88 | 37.42 | 56.82 | 65.24 | 126.19 |
| ANN | MAE | 2013.66 | 3031.77 | 3597.95 | 3776.09 | 544.88 | 982.02 | 1235.81 | 1620.53 |
| | RMSE | 2929.49 | 4484.67 | 5292.93 | 5287.87 | 907.02 | 1857.68 | 2214.92 | 2491.43 |
| | MAPE (%) | 21.07 | 28.55 | 34.53 | 42.23 | 33.2 | 53.24 | 71.76 | 106.96 |
| LSTM | MAE | 1731.31 | 2920.47 | 3423.26 | 3498.71 | 524.56 | 910.11 | 1050.78 | 1421.89 |
| | RMSE | **2510.84** | 4466.4 | 4936 | 5040.48 | 830.75 | **1465.38** | 1866.4 | 2434.48 |
| | MAPE (%) | 17.72 | 24.22 | 34.69 | 39.54 | 28.23 | 52.44 | 57.75 | 85.02 |
| Tsformer | MAE | **1547.59** | **2682.61** | **2867.71** | **3481.77** | **470.54** | **852.48** | **948.58** | **1291.55** |
| | RMSE | 2599.83 | 4282.2 | **4276.85** | **4990.61** | **779.67** | 1687.76 | **1714.13** | **2276.43** |
| | MAPE (%) | **13.77** | **23.37** | **30.69** | **33.75** | **25.51** | **41.13** | **41.22** | **64.88** |

*4.7. Ablation study*

The Tsformer compared with the baseline models in Table 6 uses the calendar of days to be forecasted. In order to evaluate the impact of using the calendar of days to be forecasted on forecasting performance, ablation studies on the Jiuzhaigou valley and Siguniang mountain datasets are conducted to compare the forecasting performance of the Tsformer with and without the calendar of days to be forecasted. Table 7 presents the results of the ablation studies. In 1-day ahead forecasting on Jiuzhaigou valley dataset, the Tsformer with calendar performs better than the Tsformer without calendar under three metrics of MAE, RMSE and MAPE. In 7-day to 30-day ahead forecasting, the Tsformer with calendar outperforms the one without calendar under at least two metrics. For the Siguniang mountain dataset, MAE of the Tsformer with calendar is slightly higher than that of the Tsformer without calendar, but the Tsformer with calendar gets better performance in other forecast horizons. The ablation studies on the two datasets demonstrate that the performance degrades when the calendar of days to be forecasted is not used. Using the calendar of days to be forecasted contributes to the forecasting performance of the proposed Tsformer. What's more, the ablation studies also indicate that even if the calendar of days to be forecasted is not used, The Tsformer without calendar still outperforms the other baseline models in Table 6 in most cases.

Table 7. Results of ablation studies on the Jiuzhaigou valley and Siguniang mountain datasets.

| Model | Metric | Dataset | | | | | | | |
|---|---|---|---|---|---|---|---|---|---|
| | | Forecast horizon | | | | | | | |
| | | Jiuzhaigou valley | | | | Siguniang mountain | | | |
| | | 1 | 7 | 15 | 30 | 1 | 7 | 15 | 30 |
| Tsformer w/ calendar | MAE | **1547.59** | **2682.61** | **2867.71** | **3481.77** | 470.54 | **852.48** | **948.58** | **1291.55** |
| | RMSE | **2599.83** | 4282.2 | **4276.85** | **4990.61** | **779.67** | **1687.76** | **1714.13** | **2276.43** |
| | MAPE (%) | **13.77** | **23.37** | 30.69 | **33.75** | **25.51** | **41.13** | **41.22** | **64.88** |
| Tsformer w/o calendar | MAE | 1604.37 | 2744.57 | 2927.15 | 3490.61 | **469.32** | 887.95 | 1011.57 | 1324.83 |
| | RMSE | 2607.03 | **4252.23** | 4359.9 | 4997.47 | 793.07 | 1800.1 | 1866.86 | 2312.94 |
| | MAPE (%) | 14.8 | 25.53 | **30.65** | 33.75 | 25.66 | 44.98 | 47.99 | 67.8 |

## 5. Interpretability

In order to recognize the main focus of the Tsformer, a study on the interpretability of the Tsformer is conducted. In the 1-day ahead forecasting using day 1 to day 7 to forecast day 8 on the Jiuzhaigou valley dataset, self-attention and encoder-decoder attention weight matrix in each decoder layer are visualized. All samples from the train set are inputted into the Tsformer. After forward propagation, the attention weight matrix of each sample is obtained. To be more generalized, the attention weights of all samples are averaged. As shown in Fig. 13, attention weights to other time steps are different for each time step. The decoder target masking mechanism makes the attention weight of the follow-up time steps of each time step be 0. It successfully prevents the Tsformer from paying attention to information after the current calculating time step, which avoids redundant information.

Fig. 13 presents the self-attention weight matrix of each decoder layer in the Tsformer. In the process of generating day 8, layer 1 pays more attention to day 7 and day 8 itself, and layer 2 mainly attends day 5, 6 and 7. Layer 3 mainly attends day 6, 7 and 8, and the closer the day to day 8 is, the larger attention weight it gets.

Layer 4 also follows this rule, which mainly attends day 4 to day 8. The weight of day 8 is close to day 7, which indicates that the Tsformer uses the information from recent days and the calendar introduced by day 8.

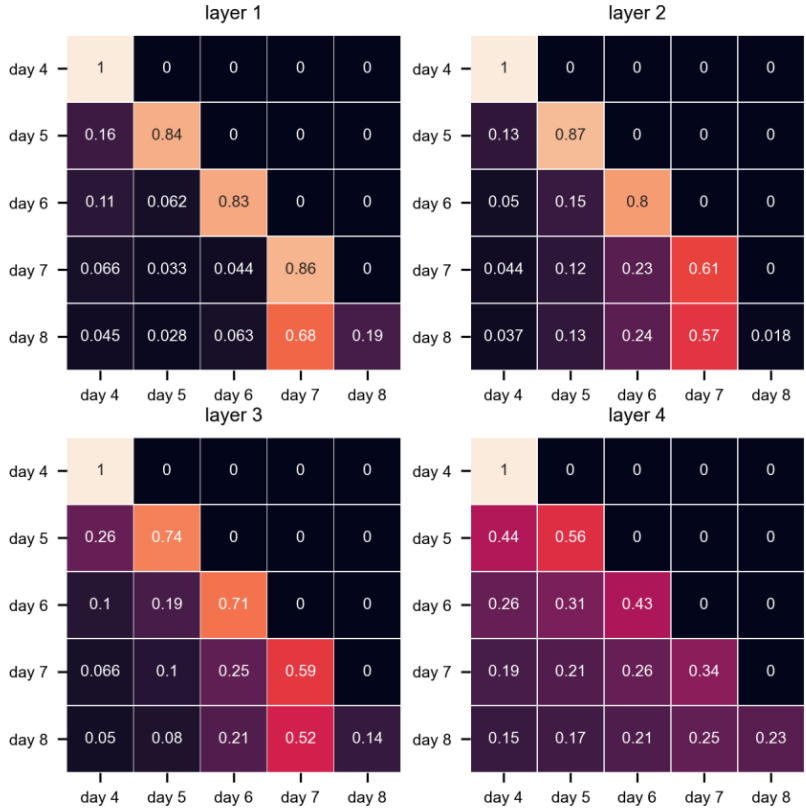

Fig. 13. Visualization of self-attention weight matrix in each decoder layer.

Fig. 14 presents the encoder-decoder attention weight matrix of each decoder layer in the Tsformer. When performing encoder-decoder attention calculation, each time step in the decoder can only attend the previous time steps in the encoder. The attention weights of other time steps are 0, which indicates that the decoder memory masking mechanism is working properly. In layer 1, day 3 to day 7 get larger attention weight. In layer 2, day 1 to day 7 get similar attention weights, but day 1 is mainly attended, which demonstrates that the Tsformer uses seasonal features in weeks in the task of forecasting tourism demand on the 8th day using the previous seven days. Day 1 to day 7 are attended in layer 3, and days closer to day 8 get larger attention weights. The visualization of the encoder-decoder attention reveals that the Tsformer tends to use information from the previous five days for 1-day ahead forecasting and can capture seasonal features of tourism demand.

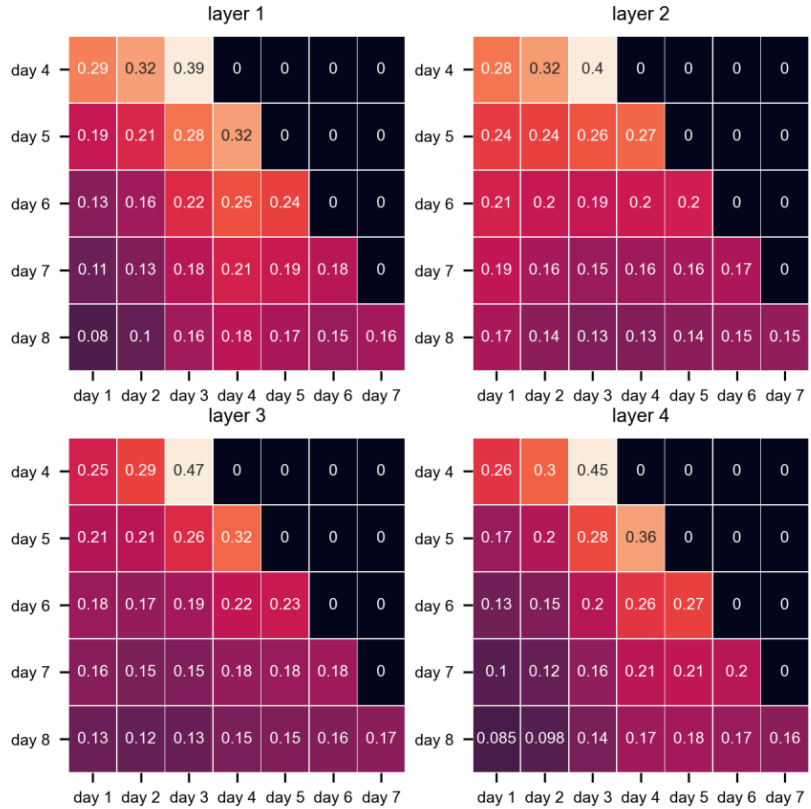

Fig. 14. Visualization of encoder-decoder attention weight matrix in each decoder layer.

## 6. Conclusion

Current AI-based methods are weak in long-term dependency capturing, and most of them lack interpretability. This study aims to improve the Transformer used for machine translation and propose the Tsformer for short-term and long-term tourism demand forecasting. In the proposed Tsformer, encoder and decoder are used to capture short-term and long-term dependency, respectively. Encoder source masking, decoder target masking, and decoder memory masking are used to simplify the interactions of attention, highlight the dominant attention, and implement time series processing in multi-head attention. For further improvement of performance, the calendar of days to be forecasted is used.

In this study, the Jiuzhaigou valley and Siguniang mountain tourism demand datasets are used to evaluate the performance of the proposed Tsformer and nine baseline models. The experiments results indicate that in the short-term and long-term tourism demand forecasting, the proposed Tsformer performs better than the baseline models under the three metrics of MAE, RMSE and MAPE in most cases and gets competitive performance. To understand the impact of using the calendar of days to be forecasted, ablation studies are conducted to evaluate the performance of the Tsformer with or without calendar. The results of ablation studies demonstrate that using the calendar of days to be forecasted contributes to the performance of the proposed Tsformer. To improve the interpretability of the Tsformer as a deep learning model, we visualized self-attention and encoder-decoder attention in each decoder layer of the Tsformer to recognize the features that the Tsformer pays attention to. The visualization indicates that the Tsformer mainly attends recent days and earlier days to capture seasonal features. Tokens that include the calendar of days to be forecasted also get large attention weights, demonstrating that the proposed Tsformer can extract rich features of tourism demand data.

Although the proposed Tsformer gets excellent performance in tourism demand forecasting, there are some deficiencies in our study. The time complexity of the proposed Tsformer is $O(n^2)$, which is the same as valina Transformer. The impact on computing efficiency is not significant on small datasets due to the parallelization characteristic of the multi-head attention mechanism, but larger datasets and longer sequences may significantly influence the computing efficiency. Therefore, in future research, further optimization of the multi-head attention mechanism that makes the Tsformer on balance between time, space complexity and forecasting performance is essential. What's more, calendar, date type and weather are discrete variables, encoding them to dummy variables may not be the perfect solution. The methods of using continuous and discrete variables are also worth discussing. Although using an embedding layer to transform discrete variables into dense vectors on small datasets may cause overfitting, it may bring additional performance on larger datasets. The features extracted by deeper networks will be more abstract; it will be more challenging to obtain interpretability by visualizing the attention weight matrix. For the more accurate recognition of complex attention dependencies and inputs that contribute to the model, methods except for visualization of attention weight matrix need to be further studied. Also, the proposed Tsformer has an excellent performance in time series forecasting and can be used in other fields in future studies.

## CRediT authorship contribution statement

**Siyuan Yi:** Conceptualization, Data curation, Investigation, Methodology, Software, Validation, Visualization, Writing - original draft. **Xing Chen:** Project administration, Resources, Supervision, Writing - review & editing, Investigation. **Chuanming Tang:** Writing - review & editing, Investigation.

## Declaration of competing interest

The authors declare that they have no known competing financial interests or personal relationships that could have appeared to influence the work reported in this paper.

## Acknowledgement

The work described in this paper was supported by Open Foundation of the Research Center for Human Geography of Tibetan Plateau and Its Eastern Slope (Chengdu University of Technology, Grant No. RWDL2021-YB001).